\documentclass[10pt,twocolumn,letterpaper]{article}

\usepackage[pagenumbers]{cvpr} 

\usepackage{graphicx}
\usepackage{amsmath}
\usepackage{amssymb}
\usepackage{booktabs}

\usepackage[pagebackref,breaklinks,colorlinks]{hyperref}

\usepackage[capitalize]{cleveref}
\crefname{section}{Sec.}{Secs.}
\Crefname{section}{Section}{Sections}
\Crefname{table}{Table}{Tables}
\crefname{table}{Tab.}{Tabs.}

\def\cvprPaperID{*****} 
\def\confName{CVPR}
\def\confYear{2022}

\usepackage{bibentry}
\usepackage{multirow}
\usepackage{pifont}
\usepackage{makecell}

\begin{document}

\title{Making Large Vision Language Models to be Good Few-shot Learners}

\author{
Fan Liu$^{1}$ \and
Wenwen Cai$^{1}$  \and
Jian Huo$^{1}$ \and
Chuanyi Zhang$^{1}$ \and
Delong Chen$^{2}$ \and
Jun Zhou$^{3}$ \and
\\
$^{1}$Hohai University\quad
$^{2}$HKUST\quad
$^{3}$Griffith University\quad
}

\maketitle

\begin{abstract}
    Few-shot classification (FSC) is a fundamental yet challenging task in computer vision that involves recognizing novel classes from limited data. While previous methods have focused on enhancing visual features or incorporating additional modalities, Large Vision Language Models (LVLMs) offer a promising alternative due to their rich knowledge and strong visual perception. However, LVLMs risk learning specific response formats rather than effectively extracting useful information from support data in FSC tasks.
    In this paper, we investigate LVLMs' performance in FSC and identify key issues such as insufficient learning and the presence of severe positional biases.
    To tackle above challenges, we adopt the meta-learning strategy to teach models ``learn to learn". By constructing a rich set of meta-tasks for instruction fine-tuning, LVLMs enhance the ability to extract information from few-shot support data for classification. 
    Additionally, we further boost LVLM's few-shot learning capabilities through label augmentation and candidate selection in the fine-tuning and inference stage, respectively. Label augmentation is implemented via a character perturbation strategy to ensure the model focuses on support information. Candidate selection leverages attribute descriptions to filter out unreliable candidates and simplify the task.
    Extensive experiments demonstrate that our approach achieves superior performance on both general and fine-grained datasets. Furthermore, our candidate selection strategy has been proved beneficial for training-free LVLMs.
\end{abstract}


\section{Introduction}

Few-shot classification (FSC), a specific application of few-shot learning (FSL)~\cite{wang2020generalizing}, draws inspiration from human learning capabilities. It enables models to classify even previously unseen classes using minimal labeled data.
Typical research focused on training robust visual embedding networks~\cite{vinyals2016matching} or leveraging additional attributes~\cite{reed2016learning} to mitigate the lack of supervision.
Nevertheless, the small amount of data tends to result in unsatisfying generalization.


Recently, large Vision Language Models (LVLMs) like GPT-4V~\cite{achiam2023gpt} and Qwen-VL~\cite{bai2023qwen} integrate powerful language models with advanced visual encoders.
They obtain rich perceptual capabilities and comprehensive knowledge via training on extensive multimodal data. They also have in-cotext learning ability that learns from demonstrations and become potentially suitable for FSC.
However, researchers~\cite{tai2023link} have found that current LVLMs can hardly link unseen image-text pairs and recognize novel categories from support samples. 
It results from that models tend to focus on specific answer formats instead of grasping the provided information.

\begin{figure}[t]
    \centering
    \includegraphics[width=1\columnwidth]{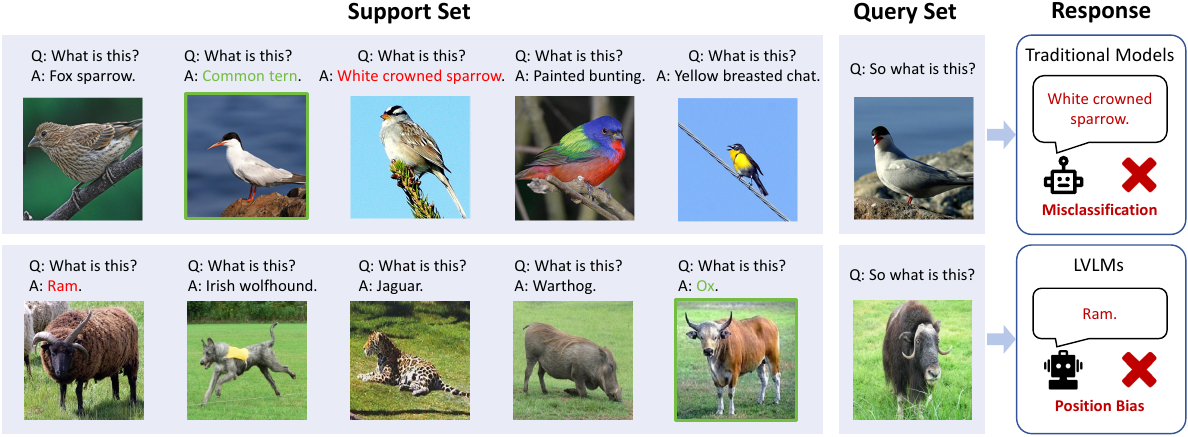}
    \caption{The challenges of FSL. Typical models often suffer from poor generalization, leading to incorrect classifications. Directly applying LVLMs to FSL also encounters positional bias that models favor the first option they encounter.}
    \label{fig:difficulty for FSL}
\end{figure}

Nevertheless, previous works did not explore the base-to-novel experimental setup in FSC, where inference classes do not overlap with training classes. 
In this paper, we attempt to evaluate the performance of LVLMs in base-to-novel few-shot setting. 
We observed that LVLMs could hardly utilize the information from support samples and achieved suboptimal classification performance. Through the examination of their outputs, we noticed a positional bias: LVLMs tend to favor the first few candidate answers they encounter.
Given the challenges LVLMs face in FSC, we adopt a meta-learning~\cite{hospedales2021meta} strategy to teach models to effectively learn from support samples. 
It is a process by which individuals increasingly manage their internalized perception, exploration, learning, and growth habits~\cite{saunders2020learning}. Meta-learning has since evolved into the concept of “learn to learn” and become a classic paradigm in FSL.

To this end, we explore the challenges and opportunities of applying LVLM to FSC.
‌Initially, we construct a rich set of instruction-following meta-tasks sourced from diverse domains. This operation enables the model to learn how to extract classification information from support data. 
To further improve the few-shot learning capabilities of LVLMs, we design label augmentation and candidate selection methods in the the fine-tuning and inference stage, respectively.
Specifically, LVLMs can sometimes be overly confident, relying on existing knowledge and neglecting information from the support data. Considering the autoregressive token modeling strategy of LVLMs, we adopt a straightforward yet effective character perturbation strategy in class names as label augmentation during fine-tuning. This strategy enhances the model's focus on the candidate classes in the support samples instead of knowledge from pre-training.
Moreover, LVLM's strong chain-of-thought (CoT)~\cite{wei2022chain} and image captioning capabilities motivate us to develop an adaptive attribute description generator to provide additional information for candidate selection. Directly entering these descriptions into LVLM may complicate the context and degrade performance. Instead, we use aggregated text similarity scores on these descriptions to select candidates. This approach filters out unreliable options and simplifies the classification task.
Finally, we conduct a thorough analysis of the initial suboptimal performance of LVLMs in FSC and why our method effectively addresses these issues and enhances performance.

Our contributions can be summarized as follows:
\begin{itemize}
\item We investigate the initial challenges LVLMs face in FSC and propose a meta-learning-based instruction fine-tuning approach. This method enhances the LVLM's ability to learn from support data in few-shot scenarios.
\item We develop two strategies to optimize LVLMs across different stages: 1) During instruction fine-tuning, we introduce label augmentation to ensure the model focuses on support data; 2) At inference, we implement an adaptive pipeline that generates and utilizes auxiliary attribute descriptions for candidate selection.
\item Our approach achieves state-of-the-art performance on eight FSC benchmarks, demonstrating the feasibility of applying LVLMs to both general and fine-grained image classification tasks. This success highlights the effectiveness of our meta-learning-based fine-tuning and semantic augmentation strategies.
\end{itemize}

\begin{figure*}
    \centering
    \includegraphics[width=\linewidth]{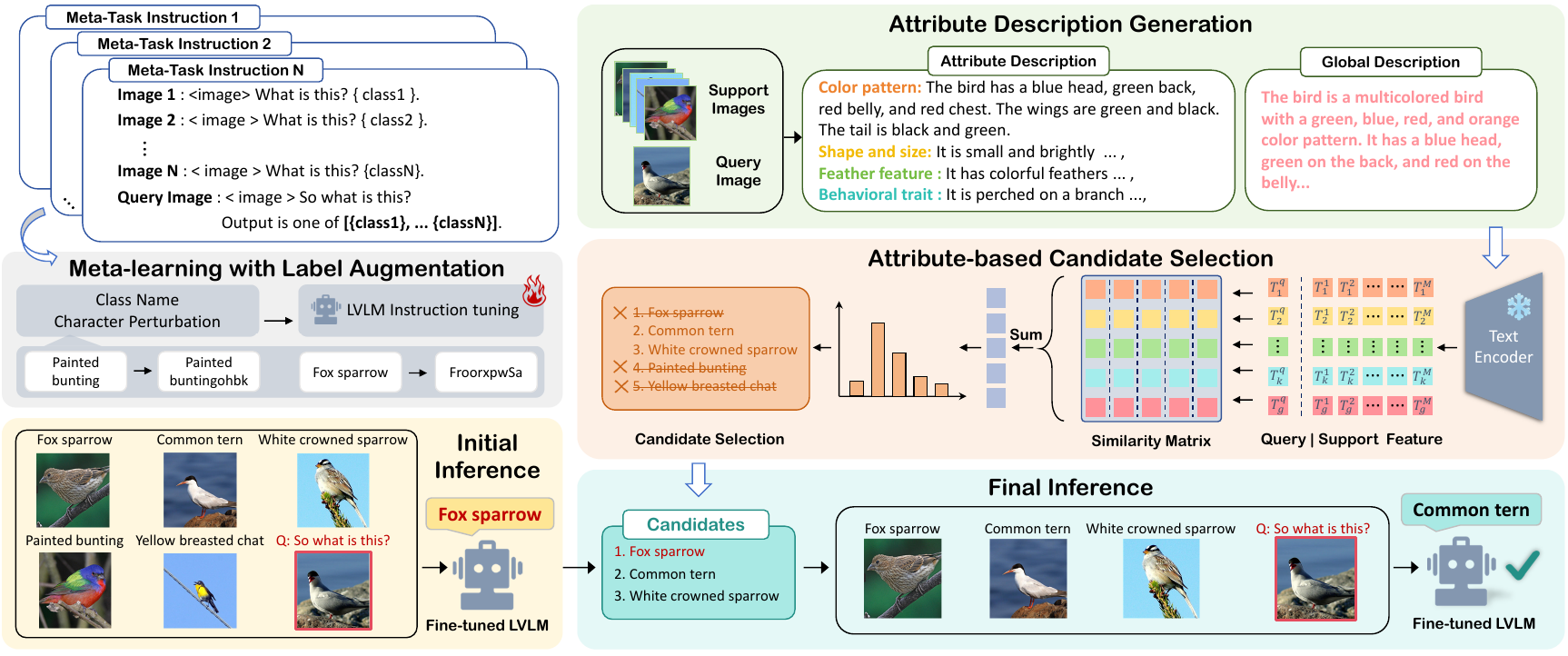}
    \caption{Overview of our approach. We construct meta-task instructions for the dataset and fine-tune LVLM using character perturbation as label augmentation. In the inference phase, we first generate attribute descriptions for each image in the meta-task instructions through an adaptive framework. Then, we leverage these descriptions to select candidate classes. If the model's initial inference does not match any of the candidates, we reorganize the meta-task instructions and query the model again for the final inference.} 
    \label{method1.jpg}
\end{figure*}

\section{Related work}
\subsection{Few-Shot Learning}
Few-shot learning aims to recognize new categories with limited labeled data. This research area can be further divided into visual-based methods and semantic-based methods. Visual-Based methods focus on extracting category-related features from images for classification. These methods can be roughly divided into two categories: optimization-based methods and metric-based methods. The former aims to learn a set of initial model parameters that can quickly adapt to new categories~\cite{finn2017model,ravi2016optimization,rusumeta}. The latter aims to learn a metric space where inter-class distances are maximized, while intra-class distances are minimized~\cite{huang2021local,snell2017prototypical,sung2018learning,vinyals2016matching}. Semantic-Based methods attempt to enhance visual recognition performance by fusing the complementary information of visual and textual modalities~\cite{chen2023semantic,li2020boosting,xing2019adaptive,xu2022generating,zhang2021prototype}. These methods usually introduce complex network frameworks to effectively utilize textual information. For example, Xing et al.~\cite{xing2019adaptive} proposed an adaptive fusion mechanism to combine visual prototypes with semantic prototypes obtained through word embeddings of class labels. Peng et al.~\cite{peng2019few} utilized graph convolutional networks to incorporate additional knowledge from knowledge graphs. Yan et al.~\cite{yan2022inferring} proposed a word embedding-guided attention mechanism to obtain label prototypes for multi-label few-shot learning problems. Different from these, our method does not require manual annotation to collect complex textual knowledge or design a complex network architecture, but instead makes full use of the rich knowledge of LVLM and its good alignment between image and text to perform classification on new classes.


\subsection{LVLM Instruction Tuning}
LVLM marks a major leap forward in vision-language modeling, which is designed to process and interpret cross-modal  information, capable of proficiently handling complex tasks that require deep understanding of context. Inspired by the remarkable success of Large Language Model (LLM) instruction fine-tuning~\cite{ouyang2022training,wang2022benchmarking}, the LVLM community has increasingly focused on incorporating instruction-following data to further enhance the model’s understanding of downstream tasks. Recently, LLaVA-1.5~\cite{liu2024visual} improves its instructions based on LLaVA's framework and obtains performance on a wider range of VQA tasks. InstructBLIP~\cite{Dai2023InstructBLIPTG} enhances zero-shot capabilities by performing instruction fine-tuning on multiple datasets. Shikra~\cite{chen2023shikra} and Kosmos-2~\cite{peng2023kosmos} extend LVLM to visual ground truth tasks using instructions with bounding box coordinates. Qwen-VL-Chat~\cite{bai2023qwen} improves the model's multi-round dialogue interaction capabilities through instruction fine-tuning.
Therefore, in this context, we introduce the meta-learning paradigm, focusing on organizing Meta-task instruction-following datasets and fine-tuning MLLM. The purpose of this approach is to fully tap the potential of MLLM in few-shot classification tasks. By leveraging meta-learning, we expect the model to better adapt to downstream tasks.

\section{Method}
\subsection{Problem Definition}

The dataset of FSL is generally divided into two parts: a base set $D_{base} = \{(x,y)\vert x\in \mathcal{X}_{base},y\in \mathcal{C}_{base}\}$ for pre-training to initialize the model and a novel set $D_{novel} = \{(x,y)\vert x\in \mathcal{X}_{novel},y\in \mathcal{C}_{novel}\}$ for testing, where $x$ denotes the image and $y$ represents the label. The label space for both sets are disjoint, meaning that $C_{base} \cap C_{novel} = \varnothing$. During testing, the support set $S = \{(x_{i}, y_{i})_{i=0}^{N \times K}\}$ is randomly selected from $D_{novel}$, which includes $N$ classes, each containing $K$ samples. The model must then accurately classify the images in the query set $Q = \{(x_{i}, y_{i})_{i=0}^{N \times M}\}$ into one of the $N$ classes present in the support set $S$, where $M$ is the number of query samples per class. This classification task is generally referred to as an $N$-way $K$-shot task.

\subsection{Instruction Tuning with Label Augmentation}

To explore the direct application of LVLM on FSL tasks, we first organized the commonly used FSL evaluation datasets into \textit{N}-way \textit{K}-shot format. To be detailed, we design Meta-Task Instructions to prompt LVLM to generate responses, as illustrated Figure~\ref{method1.jpg}.
However, we found that straightforwardly applying LVLM on $N$-way $K$-shot FSL did not yield satisfactory performance.
To enhance LVLM's performance, we adopt an instruction tuning method in a meta-learning manner.
Specifically, we collects various datasets from areas such as scene recognition, general object recognition, sentiment analysis, fine-grained recognition, and remote sensing. These fine-tuning datasets are then organized into meta-task instructions.

Following the meta-task instruction tuning, LVLM evaluates the similarity between query and support samples, and aligns query samples with candidate answers.
Consequently, the fine-tuned model can make more accurate predictions from the limited examples provided by the meta-task instructions.
However, LVLM can sometimes be overly confident, relying on categories seen in the pre-training data and overlooking the support information during query sample classification. To avoid this problem, we propose a label augmentation (LA) method via character perturbation strategy to enhance the model's focus on the support data.

Before introducing the character perturbation strategy, we should first take an inside look at the model's output process. The token embeddings $W$ of the LVLM are trained to represent the entire textual space. When given an image embedding $X_v$, the LVLM identifies the image and outputs the predicted token in the following manner~\cite{yue2024object}:
\begin{equation}
\begin{aligned}
    P(w|X_v) = argmax \ \sigma(Wf(X_v)^T) ,
\end{aligned}
\end{equation}
where $\sigma$ is the softmax function, $f$ is to transform $X_v$ for aligning with $W$, and $w$ represents the most probable single token for $X_v$.

Based on our meta-task instructions, we use interleaved $M (M=N\times K)$ image-text pairs where image features are represented as $X_v = \{X_v^i | i \in [1, M]\}$ and the prompts ``What is this? \{classname\}." are tokenized as $X_p = \{X_p^i | i \in [1, M]\}$. Query image feature is $X_v^q$, the query prompt ``So what is this? Output is one of [candidate class list] " is $X_p^q$. 
Then we can derive the complete formal expression of the input:

\begin{equation}
\begin{aligned}
    X = X_v^1 \oplus [IMG] \oplus X_p^1 \oplus ... \\
     \oplus X_v^{M} \oplus [IMG] \oplus X_p^M \\
      \oplus X_v^{q} \oplus [IMG] \oplus X_p^q,
\end{aligned}
\end{equation}
where $\oplus$ is the concatenation operation and [IMG] is a special token to indicate the boundary.
Assuming a class $C$ has $T$ tokens. Now predicting $C$ is equivalent to auto-regressively predicting its tokens:
\begin{equation}
\begin{aligned}
    P(C) = P(w_1, ..., w_T | X_b, X_p) = \prod_{t=1}^T P(w_t | w_{<t}, X),
\end{aligned}
\end{equation}
where $w_t$ is the $t$-th token of $C$, and $w_{<t}$ is the sequence of tokens before the $t$-th token.

Through the above analysis, to prevent LVLM from becoming overly confident, a straightforward approach is to disrupt common token sequences. For instance, during pre-training, the token sequence `yel'-'low' for the word `yellow' is typical. However, by perturbing the original word `yellow' to `yelowla' during fine-tuning, it results in a new token sequence `yel'-`ow'-`la'—which is strange to the LVLM. Therefore, LVLM is enforced to focus on support data instructions to learn the task paradigm.
Here is how we implemented the perturbation: 

\textbf{Split and combined:}
    split the class name according to a specific symbol such as `\underline{ }', and then re-combine the class name after splitting. \eg , `A330-300' to `300-A330'.

\textbf{Reversed:}
    take the last few characters of the class name and place them at the beginning of the other perturbation methods. \eg , `elephant' to `anteleph'.

\textbf{Random insert:}
    randomly select a number from 1 to 10 as the number of characters to randomly sample from $a$ to $z$, and insert these sampled characters into random positions within the class name. \eg ,`streetcar' to `sttrKeeutcEayrU'.

\textbf{Shuffle:}
    shuffle all characters in the class name. \eg, `shrew' to `hsewr'.

\begin{table*}[htbp]
\centering
\renewcommand\arraystretch{0.8}
\begin{minipage}[t]{0.48\textwidth}
\centering
\setlength{\tabcolsep}{1.2mm}
\begin{tabular}{ccccc}
\toprule
Method & MINI & CIFAR & TIERED & CUB \\
\midrule
BAVARDAGE~\cite{hu2023adaptive}  & 84.80  & 87.35  & 85.20  & 90.42  \\
EASY 3$\times$ResNet12~\cite{bendou2022easy}  & 84.04  & 87.16  & 84.29  & 90.56  \\
PEMnE-BMS$^{\star}$~\cite{hu2022squeezing} & 85.54  & 88.44  & 86.07 & 94.78 \\
PTMAP-SF-SOT~\cite{shalam2022self}  & 85.59  & \underline{89.94}  & -  & \underline{95.80}  \\
P\textgreater{}M\textgreater{}F~\cite{hu2022pushing} & 95.30  & 84.30  & -  & -  \\
CAML~\cite{fifty2023context}  & \underline{96.20}  & 70.80  & \underline{95.40}  & 91.80  \\
\midrule
Qwen-VL & 39.38  & 37.04  & 57.14  & 45.00  \\
Ours & \textbf{98.24}  & \textbf{95.02} & \textbf{98.06}  & \textbf{96.40}  \\
\bottomrule
\end{tabular}
\vspace{-0.2cm}
\caption{Comparison with previous work on three general datasets as well as CUB in the 5-way 1-shot setting. }
\label{table1}
\end{minipage}
\hfill
\begin{minipage}[t]{0.48\textwidth}
\centering
\setlength{\tabcolsep}{1.2mm}
\begin{tabular}{cccccc}
\toprule
Method  & CUB & Dogs & FGVC  & Flowers  & Cars  \\
\midrule
FRN~\cite{wertheimer2021few} & 83.40  & 77.53  & 87.89  & 81.22  & 87.63  \\
TDM~\cite{lee2022task} &83.25  & 76.59  & 87.91  & 82.31  & 87.69  \\
MCL-Katz~\cite{hu2023adaptive} &\underline{85.84} & 72.07  & 88.44  & 76.57  & 86.12  \\
DeepBDC~\cite{xie2022joint} &81.85 & \underline{78.81}  & 85.22  & 81.07  & 85.48  \\
LCCRN~\cite{li2023locally} &82.80  & 77.29  & 88.66  & 82.86  & 86.24  \\
SRM~\cite{li2024self} &84.14   & 77.57  & \underline{89.14}  & \underline{83.25}  & \underline{88.70}  \\
\midrule
Qwen-VL  &45.00 & 49.34  & 34.68  & 47.58 & 34.82  \\
Ours &\textbf{96.40} & \textbf{96.68} & \textbf{95.64}  & \textbf{99.58}  & \textbf{99.72}  \\
\bottomrule
\end{tabular}
\vspace{-0.2cm}
\caption{Comparison with previous work on five fine-grained datasets in the 5-way 1-shot setting.}
\label{table2}
\end{minipage}
\label{compare}
\end{table*}

\subsection{Attribute Description Generation} \label{AADG}

In the process of LVLM performing FSC, it implicitly leverages its internalized knowledge.
Since LVLM's knowledge is beneficial to FSC, we attempt to explicitly utilize it.
Considering that LVLM performs well in image captioning, we let LVLM generate image-related descriptions to assist the model inference process in subsequent sections.

Different from previous studies that only use class names or generate a single global image text description or manually select relevant attributes as additional information, we design an adaptive attribute description generation framework using LVLM to generate high-quality attributes and global descriptions for images in each category. The detailed steps are as follows:

\textbf{Step 1: Adaptive Attribute Selection.}
In this step, the type of dataset to be analyzed (\eg, bird species) and the number of attributes desired ($k$) are specified. The LVLM then suggests $k$ relevant attributes, along with brief explanations of their importance in describing the images within the specified dataset.

\textbf{Step 2: Automatic Prompt Generation.}
After obtaining attributes in step 1, LVLM is required to generate prompts for each attribute. These prompts serve as guidelines for LVLM to generate descriptions in subsequent steps. LVLM provides concise and tailored prompts for all $k$ attributes, ensuring that the generated descriptions remain focused and informative.

\textbf{Step 3: Attribute Specific Description Generation.}
For each of the $k$ attributes identified previously, the LVLM is provided with a corresponding attribute prompt from Step 2. In response, the model generates a specific detailed description of the attribute for the image.

\textbf{Step 4: Global Attribute Description Generation.}
Finally, the specific attribute descriptions from Step 3 are combined into a single, comprehensive description sentence and fed to the LVLM. The LVLM responds with its overview of the image. This attribute-global description not only captures the essence of the image, but also highlights its unique features at multiple respectives of detail.

Through our adaptive attribute description generation framework, for each support or query image in the meta-task instruction, we can obtain $k+1$ attribute descriptions regarding each image to assist the subsequent model inference process. See the appendix for more details on the attribute descriptions generation process.


\subsection{Attribute-Based Candidate Selection}

To leverage the generated attribute descriptions, we initially integrated these descriptions with meta-task instructions and then prompted LVLM. However, this method did not yield better results, as it increased the context length and introduced additional complexity.
Instead, we designed a simple yet effective semantic-based method for candidate selection (CS) using these descriptions as illustrated in Figure~\ref{method1.jpg}. This approach not only reduces the task complexity but also enhances the self-consistency of LVLM.

For each of the $M+1$ samples in the meta-task instruction, there are $k+1$ attribute descriptions. The $j-th$ description for the $i-th$ sample is denoted as $T_j^i$, where $i\in [1,M], j \in [1, k+1]$. Additionally, $T_j^q$ represents the $j-th$ description for the query sample.


We compute the text similarity $S_j \in \mathbb{R}^{1 \times {M}}$ for description $T_j$ between the query sample and support samples to obtain $k+1$ text similarity matrices. We then aggregated these similarities to obtain an overall text similarity $S_{aggr}$:
\begin{equation}
\begin{aligned}
    S_{aggr} = S_{1} + ... +S_{k+1}.
\end{aligned}
\end{equation}
Then we utilize $S_{aggr}$ to identify the top $N//2$ classes as candidate categories $C_\text{can}$, while the rest are considered unreliable.
We then compare $C_\text{can}$ with the LVLM's initial inference result $A_{ini}$. If candidate categories contains the initial inference result ($A_{ini} \in C_{can}$ ),
we consider the result to be validated. Otherwise, we reorganize the $N//2 + 1$ ways ($N//2$ ways from $C_\text{can}$ and $1$ way from $A_{ini}$) meta-task insturction to prompt LVLM again for a final inference. Final inference reduces the classification difficulty as it generates answers from fewer categories. Furthermore, this approach leverages self-consistency to enhance the reliability of the model’s output $A_{fin}$.

\section{Experiment}
\subsection{Implementation Details}

\begin{table*}[t]
    \centering
    \setlength{\tabcolsep}{1.1mm}
    \renewcommand\arraystretch{0.8}
    \begin{tabular}{cccccccccccc}
        \toprule
        \begin{tabular}{c} Meta-\\Learning \end{tabular} & 
        \begin{tabular}{c} Label \\ Augmentation \end{tabular} & 
        \begin{tabular}{c} Candidate \\ Selection \end{tabular} & 
        MINI & CIFAR & TIERED & CUB & Dogs & FGVC & Flowers & Cars & AVG($\uparrow$) \\
        \midrule
        \ding{55} & \ding{55} & \ding{55} & 39.38 & 37.04 & 57.14 & 45.00 & 49.34 & 34.68 & 47.58 & 34.82 & 43.12 \\
        \ding{55} & \ding{55} & \ding{51} & 72.30 & 68.22 & 81.66 & 59.92 & 63.52 & 65.34 & 70.70 & 68.78 & 68.81 (+25.69) \\
        \ding{51} & \ding{55} & \ding{55} & 98.20 & 95.18 & 97.96 & 93.60 & 94.82 & 93.48 & 98.32 & 99.52 & 96.39 (+53.27) \\
        \ding{51} & \ding{55} & \ding{51} & \textbf{98.28} & \textbf{95.20} & 98.00 & 94.54 & 95.40 & 94.00 & 98.72 & 99.56 & 96.71 (+53.59) \\
        \ding{51} & \ding{51} & \ding{55} & 98.24 & 94.64 & \textbf{98.06} & 96.40 & 96.40 & 95.48 & 99.38 & \textbf{99.74} & 97.29 (+54.17) \\
        \ding{51} & \ding{51} & \ding{51} & 98.24 & 95.02 & \textbf{98.06} & \textbf{96.64} & \textbf{96.68} & \textbf{95.64} & \textbf{99.58} & \textbf{99.74} & \textbf{97.45 (+54.33)} \\
        \bottomrule
    \end{tabular}
    \vspace{-0.1cm}
    \caption{Contribution of each component on eight datasets under the 5-way 1-shot setting. Meta-Learning refers to instruction tuning in a meta-learning manner. Label Augmentation means adding character perturbation during tuning. Candidate Selection means selecting more reliable candidates and re-evaluating in the inference.}
    \vspace{-0.2cm}
    \label{Ablation Study}
\end{table*}

\subsubsection{Datasets}

For instruction tuning, we selected 13 datasets from ELEVATER~\cite{li2022elevater}. These datasets span various domains such as remote sensing, scene recognition, stripe recognition, and fine-grained classification. For these datasets, we randomly split the classes into base and novel sets with a 7:3 ratio, using the base sets for fine-tuning. Note that we carefully select datasets to avoid data leakage and make fine-tuning and inference categories have no overlap.

For inference, we evaluates the proposed method on eight established FSL datasets: MiniImageNet (MINI)~\cite{vinyals2016matching}, CIFAR-FS (CIFAR)~\cite{bertinetto2019meta}, TieredImageNet (TIERED)~\cite{triantafillou2018meta}, CUB~\cite{wah2011caltech}, Stanford Dogs (Dogs)~\cite{khosla2011novel}, FGVC-Aircraft (FGVC)~\cite{maji2013fine}, Oxford 102 Flower (Flowers)~\cite{nilsback2008automated}, and Stanford Cars (Cars)~\cite{krause20133d}.
Typical FSC methods are typically tested on the first three datasets as well as CUB, while fine-grained FSC methods are evaluated on the latter five datasets.
We followed the standard base-novel split~\cite{li2019distribution}.


\subsubsection{Architecture and Training Details} \label{experiment}
We utilized the interactive Qwen-VL-Chat model as our LVLM. Its large language model was initialized with the pre-trained weights of Qwen-7B~\cite{bai2023qwen}, the visual encoder adopted the ViT architecture and was initialized with the pre-trained weights of Openclip's ViT-bigG~\cite{cherti2023reproducible}, and the visual-language adapter consisted of a single-layer cross-attention module with random initialization.

To improve training efficiency and reduce training costs, we chose the quantized version Qwen-VL-Chat-Int4 (Qwen-VL for simplicity), froze the LLM and visual encoder, and used Q-LoRA~\cite{dettmers2024qlora} to fine-tune the model's adapter. Specifically, the learning process utilized a cosine learning rate scheduler with a base learning rate of $1\times10^{-5}$ and a warm-up ratio of 0.01. Optimization was performed using the Adam optimizer, with a weight decay of 0.1 and a $\beta_2$ parameter set to 0.95, which ensured stability in convergence. The maximum sequence length of the model was set to 2048 tokens to effectively handle long sequences. Additionally, we directly used the frozen SBERT (all-MiniLM-L6-v2)~\cite{reimers2019sentence} as the text encoder used in the semantic aided inference step to measure the similarity between sentences, which had been trained on a 1B sentence pair data set and could effectively capture the semantic information of sentence vectors.

\subsubsection{Evaluation Protocol} \label{eval protocol}
Due to LVLMs' tendency to generate lengthy content and complex class names, we employ three metrics for flexible and comprehensive evaluation:

\begin{itemize}
\item $\text{Acc}$: Applying regular expression filters to remove non-alphanumeric symbols from both the candidate class names $C_\text{can}$ and LVLM outputs $A_\text{LVLM}$, then performing strict matching.
\item $\text{Acc}_\text{occur}$: After filtering, if the gold appears anywhere in $A_\text{LVLM}$, it is considered correct.
\item $\text{Acc}_\text{CLIP}$: Leveraging CLIP-L14 to map $A_\text{LVLM}$ to $C_\text{can}$ and then matching the result against gold.
\end{itemize}
Since models that are not fine-tuned often produce unreliable outputs, we use $\text{Acc}_\text{CLIP}$ as the default metric. In contrast, fine-tuned models provide more stable results, for which we use $\text{Acc}$ as the metric. The results for all three evaluation protocols will be detailed in the appendix.

\subsection{Comparison with the State-of-the-Art}
To evaluate the effectiveness of our approach, we conduct extensive experiments on eight datasets in a 5-way 1-shot setting. Table~\ref{table1} compares our results with state-of-the-art (SOTA) methods for general FSC, while Table~\ref{table2} contrasts our approach with methods specialized for fine-grained FSC.

As shown in Table~\ref{table1}, our approach outperforms existing SOTA techniques with improvements of 2.02\%, 5.08\%, 2.66\%, and 0.60\% on the MINI, CIFAR, TIERED, and CUB datasets, respectively. Moreover, our method achieves an average accuracy of 96.93\% across these four datasets, surpassing the highest average accuracy of 90.44\% achieved by PTMAP-SF-SOT by 6.49\%. It also outperforms the vision transformer-based methods P\textgreater{}M\textgreater{}F and CAML, which have average accuracies of 89.80\% and 88.55\%, respectively.
As shown in Table~\ref{table2}, in the fine-grained domain, our method improves upon SOTA techniques with gains of 10.56\%, 17.87\%, 6.50\%, 16.33\%, and 11.02\% on the CUB, Dogs, FGVC, Flowers, and Cars datasets, respectively. Our method reaches an average accuracy of 97.60\% across these five fine-grained datasets, significantly surpassing the highest average accuracy of 84.56\% achieved by the SRM method by 13.04\%.

It is evident that Qwen-VL performs poorly on both tasks. We will provide detailed analysis of this phenomenon in the Analysis Studies section.
Nevertheless, our approach can improve LVLM to achieve SOTA performance across diverse downstream datasets with just one fine-tuning, without requiring additional adjustments on the base set. This advantage indicates the superiority of applying LVLM in FSC tasks.


\begin{figure}[t]
    \centering
    \includegraphics[width=1\columnwidth]{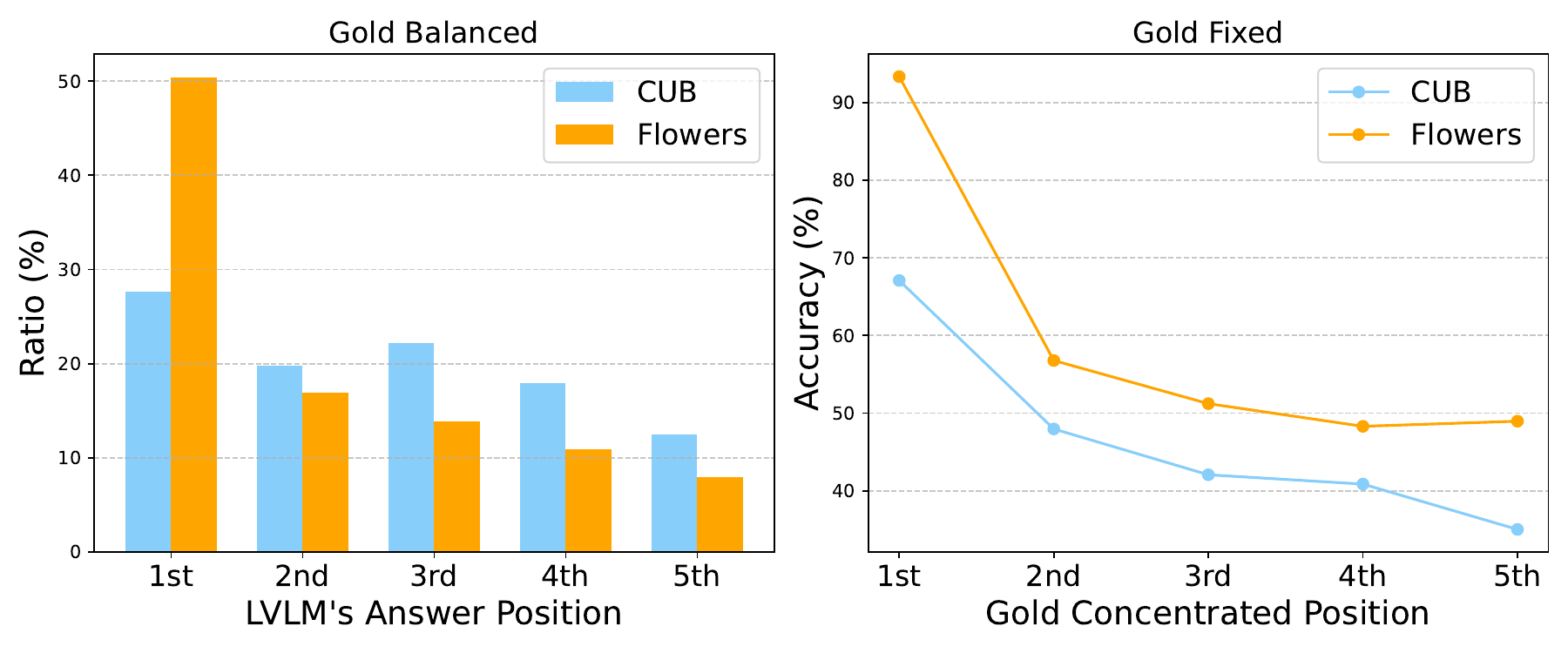}
    \caption{Illustration of position bias on CUB and Flowers under the 5-way setting: Gold Balanced means the gold answers are evenly distributed across all five candidate positions. LVLM’s Answer Position shows the actual distribution of answers provided by the LVLM. Gold Fixed indicates that gold answers are fixed in a specific position.
    Gold Concentrated Position 1st indicates that all gold answers are fixed in the first candidate position.
    }   
    \label{fig:position_cub_flower}
\end{figure}

\subsection{Analysis Studies}
\subsubsection{What is each component's contribution to the final performance?}
To validate the effectiveness of our proposed methods, we conducted ablation studies on meta-learning, label augmentation (LA), and candidate selection (CS), as detailed in Table~\ref{Ablation Study}. 
Across eight datasets, we observed the following average improvements:
1) Meta-learning fine-tuning alone led to a substantial 53.27\% improvement in model performance;
2) Adding LA, CS, and both LA and CS together resulted in performance gains of 54.17\%, 53.59\%, and 54.33\%, respectively;
3) Incorporating CS provided significant benefits to training-free methods, achieving 25.69\% accuracy gains.
These results demonstrate that each component in our proposed methods contributes to enhancing the model's few-shot classification capabilities.

\begin{figure}[t]
    \centering
    \includegraphics[width=1\columnwidth]{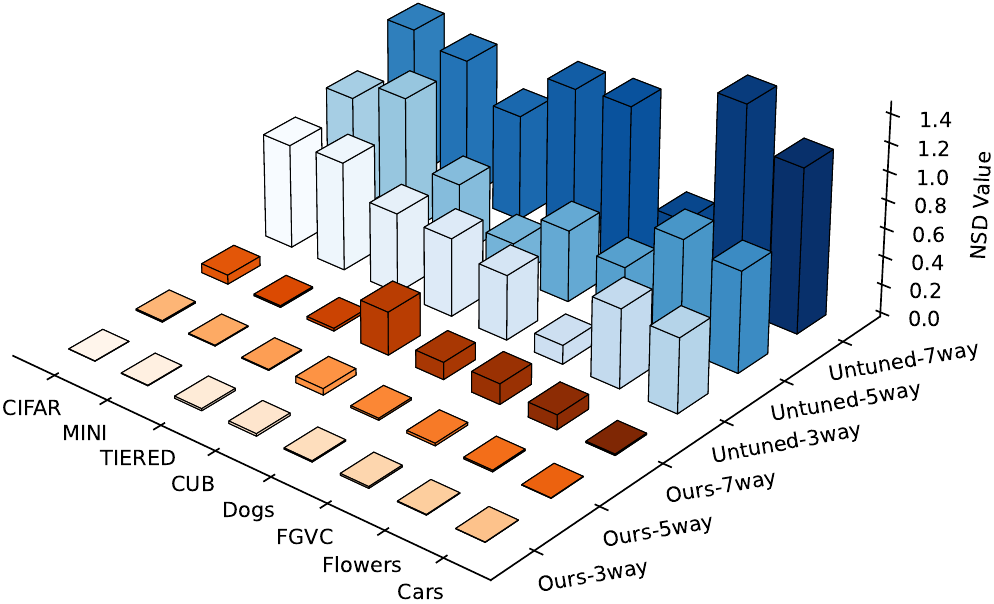}
    \vspace{-0.1cm}
    \caption{Comparison of answer position distributions between our method and Untuned LVLM. NSD values indicate the normalized standard deviation between the model's actual output positions and the uniformly sampled gold positions.} 
    \label{fig:3d}
\end{figure}

\subsubsection{Why does LVLM initially perform bad?}
We observed severe position bias in Qwen-VL across all eight datasets, with specific illustrations for CUB and Flowers shown in Figure~\ref{fig:position_cub_flower}. We compare Gold Balanced and Gold Fixed setting for detailed investigation.
In the Gold Balanced setting, gold answers are evenly distributed across the 5-way candidates, meaning they appear 1000 times at each position out of 5000 test instances. However, LVLM's actual outputs are concentrated in the early candidate positions. For example, in the Flowers dataset, more than half of LVLM’s responses selected the first candidate position.

Regarding Gold Fixed experiment setting, gold answers are in the same position among the all candidates. For example, Gold Concentrated 3rd Position means that all gold answers appear only in the third position of the candidates. When gold is concentrated in the first position, the model shows higher accuracy. However, as gold is moved to later positions, model accuracy significantly drops, indicating that the model has difficulty accessing candidates positioned farther away. 

Both experimental settings confirm that a significant reason for LVLMs' poor performance in FSC is position bias, manifested as a tendency to favor the first few candidates while having limited access to more distant ones.

Additionally, when gold is fixed in the second or third position, making it easy for the model to access, performance still remains poor. This phenomenon indicates another potential reason for LVLMs' subpar performance in FSC: their inability to effectively extract and utilize information from support samples to guide classification.

\subsubsection{Does our approach mitigate positional bias?}

We extend our experiments from the 5-way setting to the 3-way and 7-way settings, where the gold distribution is balanced. We calculate the normalized standard deviation (NSD) of the model's actual answer positions compared to the balanced gold distribution.
A higher NSD indicates a worse position bias problem where model and gold answers have greater position differences.
Figure~\ref{fig:3d} shows the NSD values for our method and the untuned LVLM across eight datasets and three \textit{N}-way \textit{K}-shot settings. The unfine-tuned LVLM exhibits severe positional bias even in the 3-way setting, which becomes more pronounced in the 7-way setting. In contrast, our model maintains balanced output distributions across 3-way, 5-way, and 7-way settings, without difficulty in accessing answers at the end of the candidate list. Notably, although our instruction-following dataset is organized for 5-way 1-shot tasks, the fine-tuned model performs well in the 7-way setting, where the candidate list is longer than that in pre-training. These results demonstrate that our method effectively mitigates the positional bias problem.

\subsubsection{Does the label augmentation strategy enhance LVLM's focus on support information?}
To further verify whether the model's performance on the inference dataset relies more on pre-trained knowledge or the provided support examples, we exchange and perturbation labels in the inference dataset.
Specifically, label exchanging makes inference data conflict with the pre-trained knowledge in LVLM, while label perturbation makes labels unseen in LVLM pre-training.

As demonstrated in Table~\ref{tab:infer_change}, the model without applying Label Augmentation (LA) does not adapt well to these conflicting or unseen scenarios. In contrast, fine-tuning with LA allows the model to better integrate information from the support examples. While the character perturbation strategy offers only slight improvements in standard FSC tasks, its effectiveness is clearly demonstrated in this experiment, enabling the LVLM to focus more on the information provided by the support examples.

\subsubsection{Why does aided semantic strategy work?}
\begin{figure}[t]
    \centering
    \includegraphics[width=1\columnwidth]{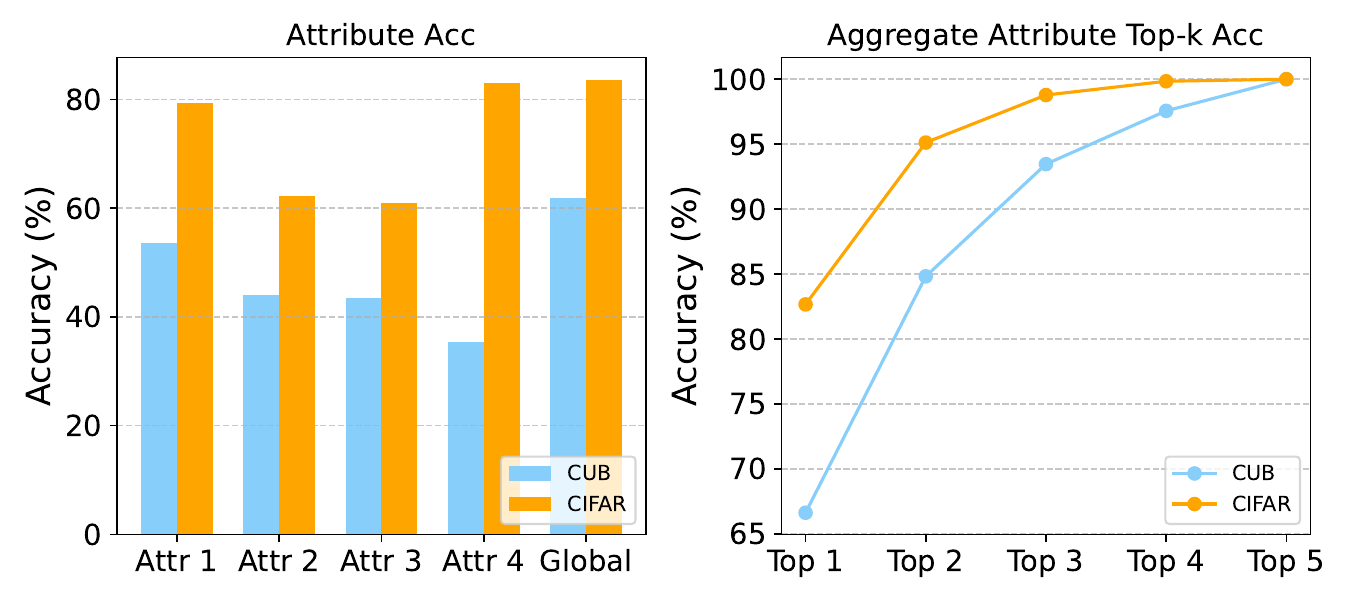}
    \caption{Left: Accuracy of each individual attribute description. Right: Top-k accuracy of aggregated attributes.}   
    \label{fig:semantic_similarity}
\end{figure}

\begin{table}[t]
    \centering
    \renewcommand\arraystretch{0.8}
    \begin{tabular}{cccc}
        \toprule
        Task       & Module       & MINI & CIFAR-FS \\
        \midrule
        \multirow{2}{*}{Label   Exchange} 
                      & w/o LA  & 17.74 & 22.60 \\
                      & w/ LA  & 40.64 & 34.04 \\
        \midrule
        \multirow{2}{*}{Label   Perturbation} 
                      & w/o LA  & 17.68 & 22.48 \\
                      & w/ LA  & 40.40 & 33.96 \\
        \bottomrule
    \end{tabular}
    \caption{Label Exchange refers to randomly reassigning class names in the dataset. Label Perturbation involves shuffling class names. Results are reported for both with and without label augmentation on these two tasks.}
    \label{tab:infer_change}
\end{table}

We computed the similarity between support samples and query samples for each individual attribute and used the maximum value as the classification result. Results for CUB and CIFAR are reported in Figure~\ref{fig:semantic_similarity} (left). These results show that a single attribute alone lacks sufficient intra-class similarity and inter-class discriminability to effectively aid LVLM classification. The performance on the CUB dataset is notably worse than on CIFAR, likely due to the finer-grained nature of the CUB dataset requiring more discriminative attribute descriptions.

As shown in Figure~\ref{fig:semantic_similarity} (right), the accuracy of aggregated attributes surpasses that of any single attribute at the top-1 level and shows significant improvement at the top-2 level. At the top-3 level, the accuracy consistently exceeds 90\%.
According to the experimental results, we utilize aggregated attributes to select $N//2$ candidates for the final reference. It is less likely to omit the gold answer and can effectively simplify the classification task in the final reference.

\begin{figure}[t]
    \centering
    \includegraphics[width=1\columnwidth]{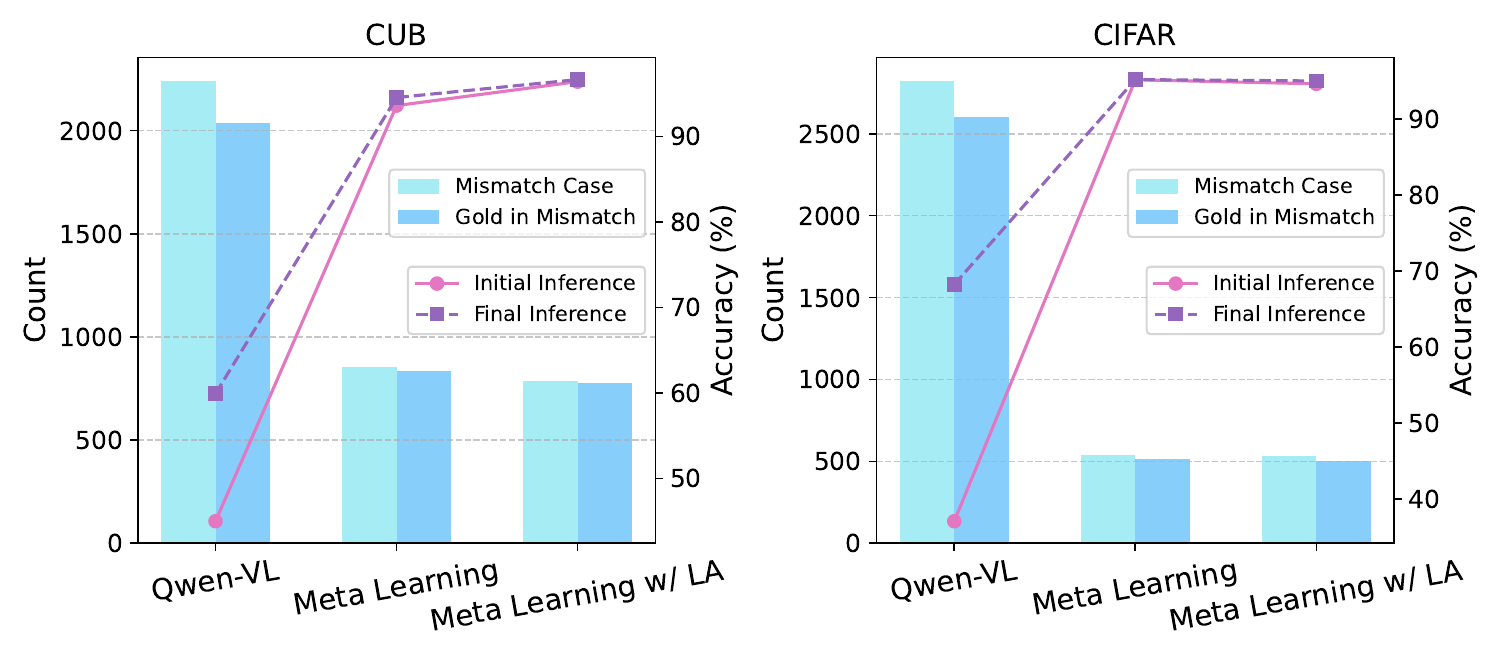}
    \caption{Candidate Selection process. Mismatch Case means the number of instances where the LVLM output does not match the top-2 attribute-based candidates. Gold in Mismatch means how many of the reorganized candidates in these mismatch cases contain the gold label. Initial Inference means the LVLM’s results before applying CS, while Final Inference shows the results after applying CS.}   
    \label{fig:filter}
\end{figure}

We also illustrate the Candidate Selection (CS) process in Figure~\ref{fig:filter}. We compare LVLM's output with the top-2 attribute-based candidates. If the LVLM's output is not among these top-2 candidates, it is considered a mismatch, and we reorganize the candidates. We count how often the gold label appears in these reorganized candidates, as this reflects the potential classification performance. The results show that the gold label appears in up to 90\% of the new candidates. We also report the accuracy of the initial inference and the accuracy after applying CS. The experiments demonstrate the effectiveness of our candidate selection method, which helps retain the correct answer and simplifies the task for better performance, especially benefiting the unfine-tuned Qwen-VL.




\section{Conclusion}
In this paper, we explored the challenges and opportunities of applying LVLM to FSC. We found untuned LVLMs could not effectively learn from support samples and suffered from position bias problems.
To enhance the few-shot learning ability of LVLMs, we organized a meta-learning-based few-shot classification instruction-following dataset. We designed label augmentation to force the model to focus more on support information in the instruction tuning. For the inference phase, we developed an attribute-based candidate selection strategy to simplify task complexity. Our method achieves state-of-the-art performance on eight general and fine-grained few-shot classification benchmarks.

{\small
\bibliographystyle{ieee_fullname}
\bibliography{egbib}
}

\end{document}